\pdfoutput=1

\documentclass[11pt]{article}

\usepackage[final]{acl}

\usepackage{times}
\usepackage{latexsym}

\usepackage[T1]{fontenc}

\usepackage[utf8]{inputenc}

\usepackage{microtype}

\usepackage{inconsolata}

\definecolor{kellygreen}{rgb}{0.3, 0.73, 0.09}
\definecolor{alizarin}{rgb}{0.82, 0.1, 0.26}

\newcommand{\ours}[1]{\textbf{D.Va}}

\usepackage{graphicx}
\usepackage{subcaption}
\usepackage{amsmath}
\usepackage{booktabs}
\usepackage{multirow}
\usepackage{makecell}
\usepackage{caption}
\usepackage{subcaption}
\usepackage{hyperref}
\usepackage{pifont}

\title{D.Va: Validate Your Demonstration First Before You Use It}

\author{Qi Zhang,~Zhiqing Xiao,~Ruixuan Xiao,~Lirong Gao,~Junbo Zhao
\thanks{Corresponding author.}
\\ Zhejiang University \\ \texttt{\{cheung\_se,zhiqing.xiao,xiaoruixuan,gaolirong,~j.zhao\}@zju.edu.cn} }

\begin{document}
\maketitle
\begin{abstract}
    In-context learning (ICL) has demonstrated significant potential in enhancing the capabilities of large language models (LLMs) during inference. It's well-established that ICL heavily relies on selecting effective demonstrations to generate outputs that better align with the expected results.
    As for demonstration selection, previous approaches have typically relied on intuitive metrics to evaluate the effectiveness of demonstrations, which often results in limited robustness and poor cross-model generalization capabilities.
    To tackle these challenges, we propose a novel method, \textbf{D}emonstration \textbf{VA}lidation (\textbf{D.Va}), which integrates a demonstration validation perspective into this field.
    By introducing the demonstration validation mechanism, our method effectively identifies demonstrations that are both effective and highly generalizable.
    \textbf{D.Va} surpasses all existing demonstration selection techniques across both natural language understanding (NLU) and natural language generation (NLG) tasks. Additionally, we demonstrate the robustness and generalizability of our approach across various language models with different retrieval models.
\end{abstract}

\section{Introduction}

    Large language models (LLMs) demonstrate impressive generalization capabilities under the in-context learning (ICL) paradigm, adapting to new tasks without parameter updates~\cite{gpt3}. 
    In this ICL setup, LLMs utilize demonstration samples provided in the input context as exemplars to guide their output generation. 
    This emergent ICL ability allows LLMs to generalize cost-effectively to unseen tasks. 
    However, prior research highlights that the quality of demonstration samples significantly impacts ICL performance~\cite{liu-etal-2022-makes,lu-etal-2022-fantastically}. 
    Poorly constructed demonstrations can significantly degrade overall performance, making effective demonstration selection a crucial area of study~\cite{iter-etal-2023-context}.

    Effective demonstration selection has become a key focus in ICL research~\cite{dong-etal-2024-survey,luo2024incontext}. 
    While early corpus-level methods relied on a fixed set of demonstrations~\cite{gpt3,shin-etal-2020-autoprompt,gao-etal-2021-making,jiang-etal-2021-know,sorensen-etal-2022-information}, recent studies emphasize dynamically selecting the most suitable demonstrations for each test input~\cite{luo2024incontext}.
    These methods fall into two categories: data-dependent strategies and self-adaptive strategies.
    Data-dependent strategies typically rely on measures, \emph{i.e.}, the textual or semantic similarity between the test input and demonstrations to 
    conduct demonstration selection.
    Such measures are often extracted by off-the-shelf retrievers such as BM25~\cite{bm25} and Sentence-BERT~\cite{reimers-gurevych-2019-sentence}.    
    Despite the simplicity, such approaches entirely hinge on a static, offline retriever, limiting its ability to generalize to previously untrained fields.    

 \begin{figure}[t]
      \centering
      \includegraphics[width=0.95\columnwidth]{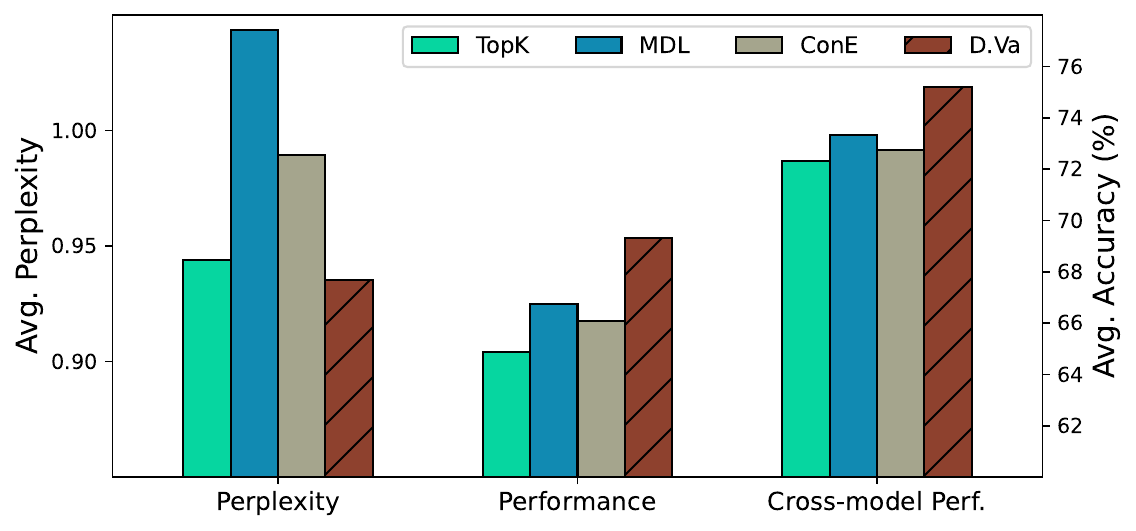}
      \caption{Collaborative comparison of the average perplexity, performance, and cross-model performance of different methods across eight NLU datasets on Llama-3.2-1B. Cross-model refers to selecting demonstrations with Llama-3.2-1B while inferring with Llama-3.1-8B. Although MDL and ConE outperform the data-dependent baseline TopK in terms of performance, they don't effectively reduce the model's perplexity on the ground-truth labels and show limited cross-model generalization capabilities.}
      \label{fig:ppl_vis}
    \end{figure}

    Another research line in this field is self-adaptive strategies, adopting a more dynamic way.
    \citeauthor{wu-etal-2023-self} proposed a model-based metric to evaluate demonstration effectiveness from a novel self-adaptive view of demonstration selection. They introduced a select-then-rank framework that leverages off-the-shelf retrievers to retrieve a candidate set and re-rank them based on their metric.
    Later,~\citeauthor{peng-etal-2024-revisiting} further demonstrated that the language model's understanding of the test input could help identify effective demonstrations.
    These methods typically retrieve a set of candidate demonstrations and then select the most effective ones based on their proposed metrics.

    Self-adaptive methods, while often outperforming data-dependent ones, still face significant challenges.
    Due to their dependence on superficial metrics for selection, these adaptive methods can exhibit subpar performance when applied in cross-model and other real-world scenarios, sometimes even yielding worse results than basic data-dependent methods
    ~\cite{dong-etal-2024-survey}.
    Through extensive observations and analysis, we conclude that these shortcomings stem from their inability to fully capture the fundamental essence of demonstration selection in ICL.
    The key challenge lies in identifying demonstrations that can effectively guide the language model to generate the ground-truth answer with minimal perplexity.
    However, the absence of ground-truth labels during the selection process makes it inherently difficult to evaluate the quality of demonstrations from this perspective.

    To address these challenges, we introduce \textbf{D}emonstration \textbf{VA}lidation (\ours{}), a novel self-adaptive demonstration selection method that adopts a validation-driven perspective.
    Inspired by previous corpus-level methods~\cite{lu-etal-2022-fantastically,sorensen-etal-2022-information} that partition a separate validation set to construct a fixed demonstration set,  we intend to adapt this validation paradigm for a self-adaptive framework. Our principle is to select demonstrations via a simulated validation process, ensuring the LLM achieves minimal perplexity for the potential unseen ground-truth answer.
    However, unlike corpus-level scenarios, the distribution shift between single validation input and single test input significantly impacts the overall performance.
    To further address this challenge, we propose a preference-based calibration mechanism that adjusts the validation loss based on the language model's preferences between the test and validation inputs, effectively mitigating this phenomenon.
    As illustrated in Figure~\ref{fig:ppl_vis}, \ours{} resolves the accuracy-confidence discrepancy seen in prior methods and demonstrates strong cross-model capabilities. In general, \ours{} achieves superior, generalizable performance across diverse datasets, surpassing all existing demonstration selection methods for both natural language understanding (NLU) and natural language generation (NLG) tasks.
    
    To sum up, our contributions can be concluded as follows:
    \begin{itemize}
        \item To our best knowledge, we are the first to propose a novel demonstration validation mechanism for self-adaptive selection methods.

        \item We propose a novel demonstration selection method (\textbf{D.Va}) for in-context learning,
        which helps diverse language models achieve state-of-the-art performance on various datasets with different retrieval models.
        
        \item 
        By using small language models as surrogates for LLMs, the strong cross-model generalization capabilities of \ours{} highlight its potential in demonstration selection scenarios.
    \end{itemize}

    \section{Related Work}
    \subsection{In-Context Learning}
    It was discovered that pre-trained LLMs have remarkable capabilities in adapting to new tasks by providing a related context or several demonstrations alongside the test input~\citep{gpt3,dong-etal-2024-survey,luo2024incontext}, which is typically referred to as the in-context learning ability of LLMs. However, it's evident that the selection and order of demonstrations can largely affect the final performance~\cite{liu-etal-2022-makes,lu-etal-2022-fantastically}.

    \subsection{Demonstration Selection in ICL}
     While early corpus-level methods relied on a fixed set of demonstrations~\cite{gpt3,shin-etal-2020-autoprompt,gao-etal-2021-making,jiang-etal-2021-know,sorensen-etal-2022-information}, recent studies emphasize dynamically selecting the most suitable demonstrations for each test input~\cite{luo2024incontext}, which can be categorized into two groups: data-dependent strategies and self-adaptive strategies.

    \begin{figure*}[h]
        \centering
        \includegraphics[width=0.93\textwidth]{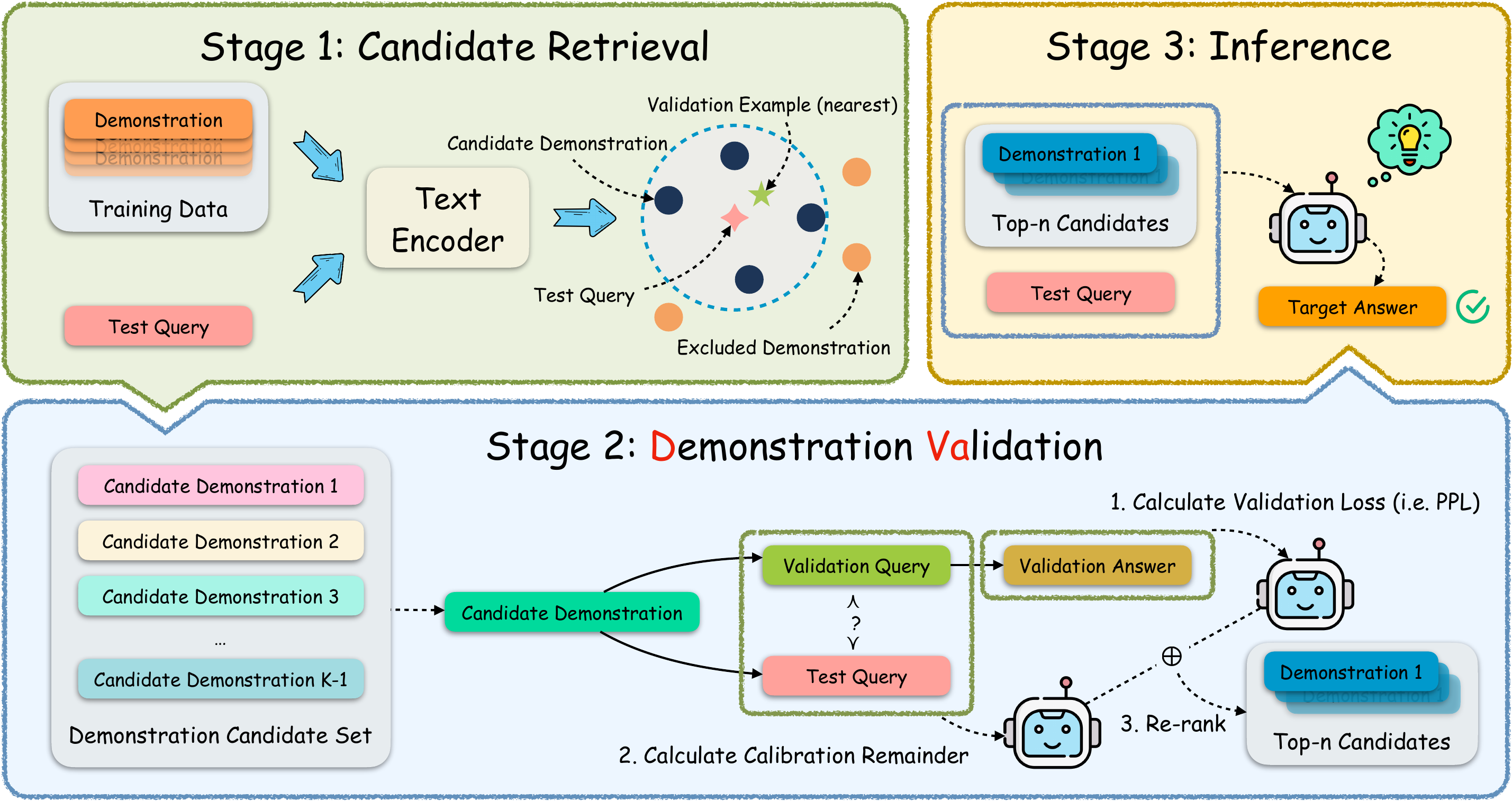}
        \caption{The main framework of \textbf{D.Va}. We first retrieve the nearest demonstration as the validation example and a demonstration candidate set of size $K-1$. Then use our proposed metric to re-rank all the candidates and concatenate the top $n$ candidates as the final context at the inference stage.}
        \label{fig:main}
    \end{figure*}

    As for data-dependent strategies, previous work always relies on the textual or semantical similarity between the test input and the demonstrations to select the most suitable demonstrations, namely retrieval-based ICL (Ret-ICL). In this circumstance, BM25~\citep{bm25} and Sentence-BERT~\citep{reimers-gurevych-2019-sentence} are commonly used to retrieve the most similar demonstrations for each test input. Besides, many researchers also focus on extracting high-quality training data and further optimizing the ability of retrievers~\cite{ceil,li-etal-2023-unified,luo2023dricldemonstrationretrievedincontextlearning,wang-etal-2024-learning}.

    In the realm of self-adaptive strategies, \citet{wu-etal-2023-self} pioneered this area by introducing a self-adaptive method for selecting effective demonstrations for classification tasks.
    Subsequently, \citet{peng-etal-2024-revisiting} leveraged the language models' understanding of test inputs together with candidate examples to identify demonstrations that effectively minimize the perplexity of the test inputs.


    \section{Methodology}
    
    In this section, we first introduce the problem formulation of in-context learning and then present our proposed method \textbf{D.Va} in detail. Figure~\ref{fig:main} briefly illustrates the framework of our method. 

    \subsection{Problem Formulation}
    The primary objective of ICL demonstration selection is to increase the probability that the model outputs the correct answer based on the selected demonstrations and test input.
    Thus, considering an ideal scenario where the ground-truth answer to the test input is available, we can find the optimal demonstration $d^{\ast}$ for ICL by solving the following optimization problem:
    \begin{equation}
        \label{eq:icl_condition}
        \begin{aligned}
            d^{\ast}=\mathop{\arg\max}\limits_{d\in D^K}P_{\theta}(y_{t}|d,x_t)
        \end{aligned}
    \end{equation}
    where $d$ and $D^K$, represent a candidate demonstration and a candidate set of size $K$. $x_t$ and $y_t$ refer to the test input and its ground truth respectively.

    Although the approach of directly selecting demonstrations based on the test input-output is intuitive and effective, in practical scenarios, the ground-truth label $y_t$ of the test input $x_t$ is unseen while inferring.
    Previous approaches can be broadly categorized into two types: one focuses on calculating the information entropy of the model under a given label space~\cite{wu-etal-2023-self}, while the other uses the model's understanding of the test inputs as the selection criteria~\cite{peng-etal-2024-revisiting}. The former entails high computational costs and is unsuitable for open-ended tasks, while the latter suffers from limited performance due to the inability to accurately measure the model's perplexity on test samples. 

    To address this challenge, we take inspiration from~\citeauthor{lu-etal-2022-fantastically,sorensen-etal-2022-information} who perform corpus-level selection with a validation set. In practice, we incorporate a query-specific validation example as the anchor instead of a validation set for the whole train set, thereby proposing a demonstration validation metric that enables indirect estimation of the model's perplexity on the unseen ground-truth labels.

    \subsection{Demonstration Validation Process}

    In this section, we propose a novel demonstration validation metric aiming to indirectly estimate the model's perplexity on the unseen ground-truth labels.
    To enable more accurate estimation, the design principle of this metric is to reflect the ability of the demonstration to guide the model in generating the ground truth.
    The core issue is twofold: i)-how to select the validation example; ii)-how to design the metric based on the validation example.
    
    We start by retrieving the nearest $k$ demonstrations as the original candidate set $D$. Then we choose the semantically nearest demonstration as the validation example to minimize the distribution shift between the validation and test example\footnote{The impact of validation example selection will be further discussed in Appendix~\ref{sec:valid_example}.}, and the remaining $K-1$ demonstration examples consist of the under-selected candidates set $D^{\prime}$:
    \begin{equation}
        \label{eq:valid-example}
        \begin{aligned}
            d_{v} & = \mathop{\arg\max}\limits_{d\in D}{\rm sim}(d,x_t)\\
            D^{\prime} & = D\backslash \{d_{v}\}
        \end{aligned}
    \end{equation}
    Since $d_v$ is the semantically nearest demonstration to the test input $x_t$, the perplexity of the language model on the validation answer $\mathcal{L}_{v}$, can be regarded as a surrogate indicator that reflects the appropriateness of a candidate $d$ as the demonstration of $x_t$ to some extent. In the scope of ICL, $\mathcal{L}_{v}$ and the target test perplexity $\mathcal{L}_{t}$ can be denoted as $\mathcal{L}_{v}=-\log P_{\theta}(y_{v}|d,x_{v})$ and $\mathcal{L}_{t}=-\log P_{\theta}(y_{t}|d,x_{t})$ respectively.
    Although a smaller $L_v$ may highlight the superiority of $d$ in assisting the model in addressing problems similar to $x_t$, this does not convincingly demonstrate that $d$ is an appropriate demonstration for $x_t$, given the distribution shift between $x_t$ and $x_v$ may be significant.
    
    To address this issue, a calibration remainder should be introduced to further approximate $\mathcal{L}_{t}$ and reduce this discrepancy. Specifically, a compensation should be applied to $\mathcal{L}_{v}$ if it is overestimated; otherwise, a penalty should be introduced if $\mathcal{L}_{v}$ is underestimated.


    Furthermore, a more intuitive challenge lies in determining whether the estimation of test perplexity $\mathcal{L}_{t}$ based on validation perplexity $\mathcal{L}_{v}$ constitutes an overestimation or an underestimation.
    Previous research has revealed a negative correlation between the language model's perplexity on the prompt and the probability of the language model correctly answering the question~\cite{gonen-etal-2023-demystifying,peng-etal-2024-revisiting}.
    Inspired by this, we further state that with a given demonstration $d$, the language model can better solve the problem that it can better understand.
    We hereby introduce a calibration remainder $\epsilon$ which helps distinguish the difference in the model's understanding of $x_{t}$ and $x_{v}$, namely, whether the estimation of $L_t$ is an overestimation or an underestimation according to its sign:
    \begin{equation}
      \label{eq:epsilon}
      \begin{aligned}
        \epsilon & = -\log \frac{P_{\theta}(x_{t}|d)}{P_{\theta}(x_{v}|d)}
      \end{aligned}
    \end{equation}

    To better integrate the two indicators above, a tunable hyper-parameter $\lambda$ is introduced to balance $\mathcal{L}_v$ and $\epsilon$. Given a demonstration $d$ and a test input $x_t$, we finally present the expression of the demonstration validation metric, denoted as ${\rm Score} (d,x_t)$:
    \begin{equation}
        \label{eq:formulation}
        \begin{aligned}
            {\rm Score} (d,x_t) = (1-\lambda)\cdot\mathcal{L}_{v}+\lambda\cdot\epsilon
        \end{aligned}
    \end{equation}

    \subsection{Overall Selection Framework}
    We further demonstrate the overall demonstration selection process based on our proposed demonstration validation metric in this section.

    Following the select-then-rank framework proposed by~\citeauthor{wu-etal-2023-self}, we first retrieve $K$ candidates for each test input due to computational consideration\footnote{In practice, we set $K=30$ for all experiments including this hyper-parameter following~\citeauthor{wu-etal-2023-self}.}. Then we adapt the semantically nearest one as the validation example, and sample the minimal-$n$ demonstrations $D_s$ (under an $n$-shot setting) according to our proposed metric:
    \begin{equation}
        \label{eq:select}
        \begin{aligned}
            D_s=\mathop{\arg \rm{sort}}_{d\in D^{\prime}}{\rm Score} (d,x_t)[:n]
        \end{aligned}
    \end{equation}

    The selected $n$ demonstrations will then be concatenated in a descending order to generate the final demonstration organization according to their corresponding scores following~\citeauthor{liu-etal-2022-makes}. The impact of demonstrations concatenation order is further analyzed in Appendix~\ref{sec:demons_order}.

    \subsection{Interpretation of $\epsilon$ from a Preference Perspective}
    When we rethink the role of $\epsilon$ in Equation~\ref{eq:epsilon}, we find that $\epsilon$ can be regarded as a transformation of the preference exhibited by the language model between two inputs. We here use Bradley-Terry (BT) model~\citep{bt_model} to describe the preference of the language model over $x_t$ and $x_v$ with demonstration $d$ as the context:
    \begin{equation}
        \label{eq:bt_preference}
        \begin{aligned}
            P_{\theta}(x_{v}\prec x_t|d)=\frac{P_{\theta}(x_t|d)}{P_{\theta}(x_t|d)+P_{\theta}(x_v|d)}
        \end{aligned}
    \end{equation}
    Thus the original expression can be transformed as:
    \begin{equation}
        \label{eq:bt_preference}
        \begin{aligned}
            \epsilon & = -\log \frac{P_{\theta}(x_{v}\prec x_t|d)}{1-P_{\theta}(x_{v}\prec x_t|d)}
        \end{aligned}
    \end{equation}
    From this viewpoint, the calibration remainder $\epsilon$ implies the language model's preference over the two queries. This perspective corroborates the phenomenon found by~\citeauthor{jiang-etal-2023-generative} where the language model tends to be more confident in answering its self-generated next prompt.





    \section{Experimental Settings}
    \subsection{Datasets}
    For natural language understanding (NLU) tasks, we evaluate our method on 8 datasets including two topic classification datasets Trec~\citep{hovy-etal-2001-toward} and AgNews~\citep{ag_news}, one multi-choice question answering dataset: Commonsense QA (CMS QA)~\citep{talmor-etal-2019-commonsenseqa}, two sentiment classification~\citep{socher-etal-2013-recursive} datasets SST-2 and SST-5, three natural language inference datasets: SNLI~\citep{bowman-etal-2015-large}, MNLI~\citep{williams-etal-2018-broad} and QNLI~\citep{wang-etal-2018-glue} following the settings of~\citet{wu-etal-2023-self}. The detailed evaluation strategies are listed in Appendix~\ref{sec:eval}.

    Besides, we also consider several natural language generation (NLG) tasks including two translation tasks from Flores200~\citep{flores2,flores1,flores0}, one question answering task SQuAD v2~\citep{rajpurkar-etal-2016-squad,rajpurkar-etal-2018-know} and one text summarization task Samsum~\citep{gliwa-etal-2019-samsum}.

    \subsection{Baselines}
    To compare \textbf{D.Va} with previous methods, we mainly take the following methods into consideration.
    \textbf{0-shot}: the zero-shot setting where no demonstration is provided. \textbf{Random}: we randomly select demonstrations from the training set. \textbf{BM25}: we use BM25~\citep{bm25} to retrieve the most similar demonstrations for each test example. \textbf{TopK}: we use Sentence-Transformer to retrieve the most similar demonstrations for each test example. \textbf{MDL}~\citep{wu-etal-2023-self}: a two-stage method that integrates Minimum Description Length (MDL) principle to demonstration selection\footnote{In this paper, MDL is only evaluated in classification tasks due to its limitation.}. \textbf{ConE}~\citep{peng-etal-2024-revisiting}: a self-adaptive demonstration selection method aims at selecting the demonstrations that can help language models understand the test input to the greatest extent.

    \subsection{Implementation Details}
    We conduct our experiments with GPT2-XL (1.5B)~\cite{radford2019language} and Llama-3 series~\cite{grattafiori2024llama3herdmodels} including Llama-3.1-8B, Llama-3.2-1B, and Llama-3.2-3B. We perform each experiment three times using different random seeds and report the average performance. Unless otherwise specified, all experiments in this paper are conducted using an 8-shot setting.
    
    For the choice of $\lambda$, we conduct exploring experiments by randomly selecting $1000$ examples as the validation set on the Trec dataset. We split the range of $\lambda$ into 10 intervals from $0.0$ to $1.0$ and the validation accuracy peaks when $\lambda=0.6$. Thus, the coefficient $\lambda$ is set to $0.6$ for all models and datasets in this paper.

    \section{Experiments}
    \subsection{Main Results}
    The main results of our method compared to other methods on Llama-3.2-1B and Llama-3.1-8B are shown in Table~\ref{tab:main-results} and Table~\ref{tab:nlg-results}. In general, our proposed method \textbf{D.Va} consistently outperforms other methods across all datasets and tasks, demonstrating the effectiveness of our method in selecting the most suitable demonstrations for in-context learning. Specifically, \textbf{D.Va} achieves an average improvement of $2.60\%$ and $0.94\%$ over the second-best method on Llama-3.2-1B and Llama-3.1-8B respectively.

\begin{figure*}[h]
    \begin{subfigure}{0.33\textwidth}
    \includegraphics[width=\linewidth]{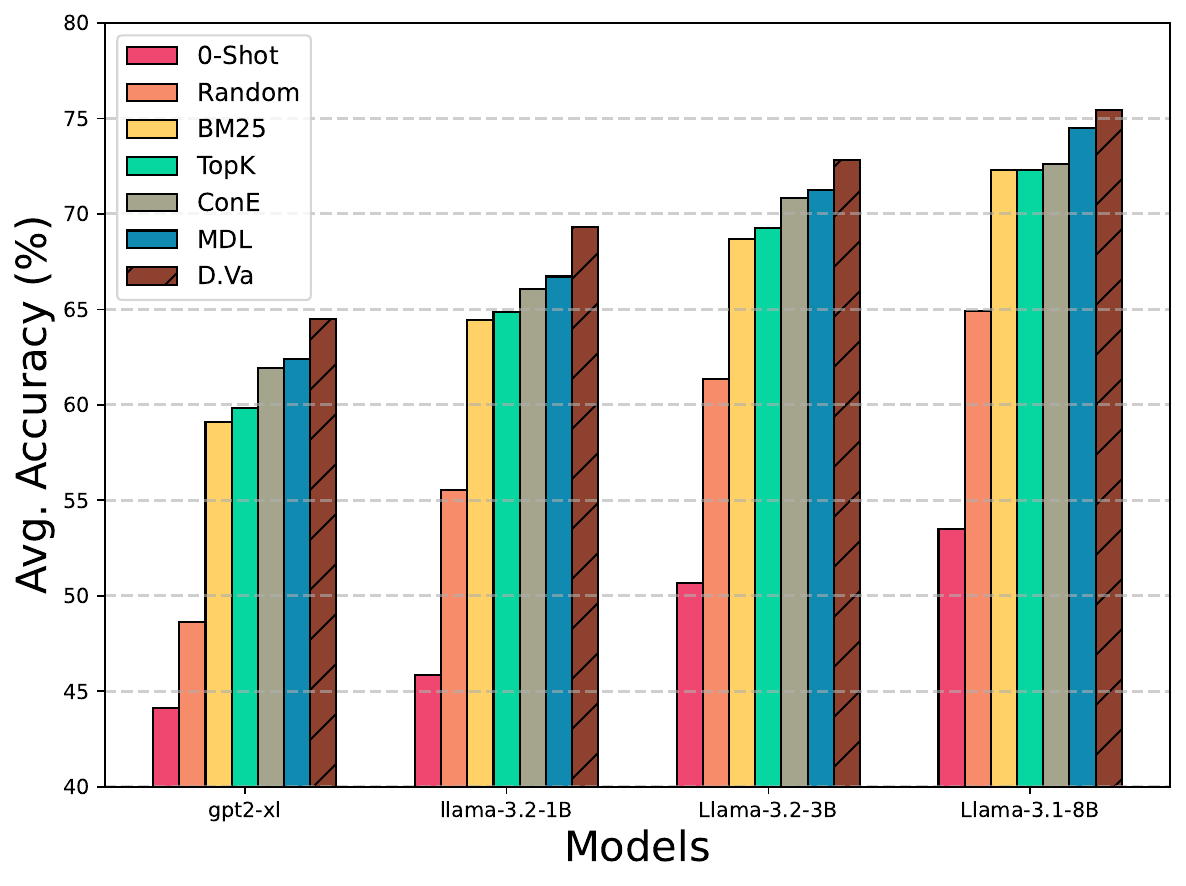} 
    \caption{}
    \label{fig:model_comparison}
    \end{subfigure}
    \begin{subfigure}{0.32\textwidth}
    \includegraphics[width=\linewidth]{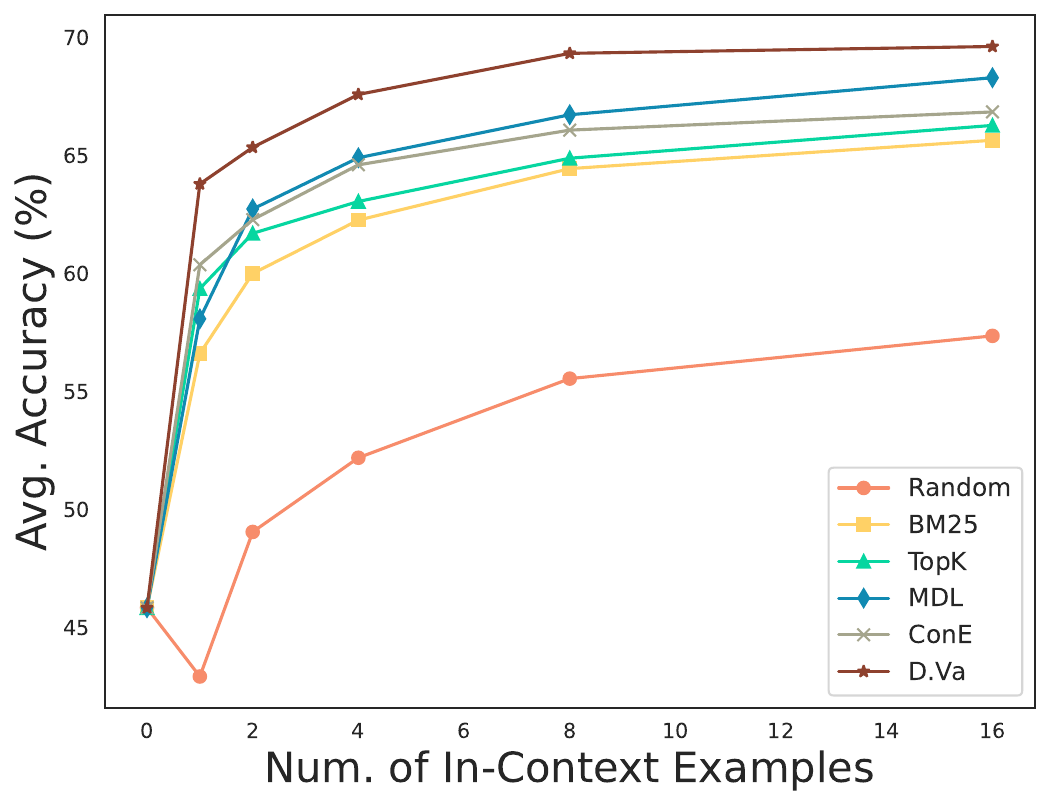}
    \caption{}
    \label{fig:in-context-example}
    \end{subfigure}
    \begin{subfigure}{0.32\textwidth}
    \includegraphics[width=\linewidth]{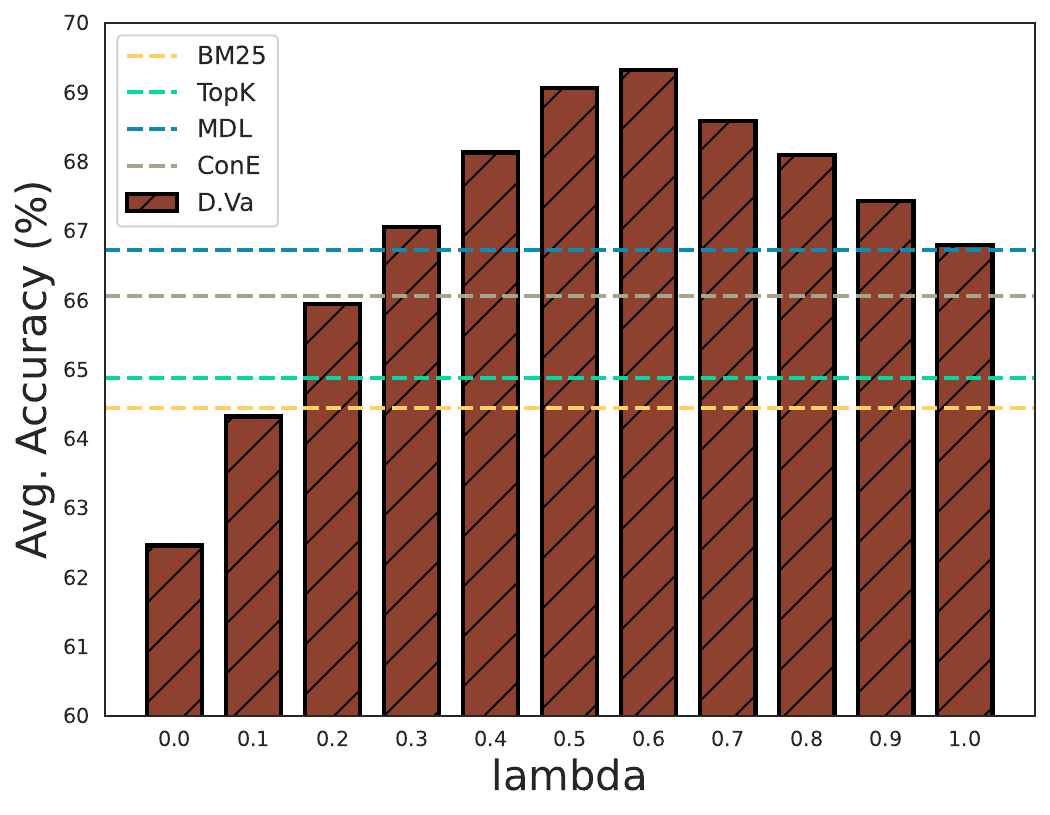}
    \caption{}
    \label{fig:lambda}
    \end{subfigure}
    \caption{(a) The performance of our method compared to other methods on GPT2-XL, Llama-3.2-1B, Llama-3.2-3B and Llama-3.1-8B, respectively. (b) The performance of various methods using different numbers of in-context examples on Llama-3.2-1B. (c) The overall performance of our method across eight NLU datasets using different values of $\lambda$ on Llama-3.2-1B.}
    \label{fig:all-fig}
    \end{figure*}

    To further show the superiority of our method compared to previous methods, we further evaluate the performance of our method on more language models including Llama-3.2-3B and GPT2-XL.
    As shown in Figure~\ref{fig:model_comparison}, our method achieves a relative improvement of $3.35\%$, $3.90\%$, $2.19\%$ and $1.26\%$ over the second-best method on GPT2-XL, Llama-3.2-1B, Llama-3.2-3B and Llama-3.1-8B respectively, which indicates that weaker language models can benefit more from our proposed method.

    \begin{table*}[h]
        \centering
        \resizebox{0.9\textwidth}{!}{%
        \begin{tabular}{@{}c|c|cccccccc|c@{}}
        \toprule
        \textbf{Model} &
          \textbf{Method} &
          \textbf{CMS QA} &
          \textbf{Trec} &
          \textbf{AgNews} &
          \textbf{SST-2} &
          \textbf{SST-5} &
          \textbf{QNLI} &
          \textbf{SNLI} &
          \textbf{MNLI} &
          \textbf{Avg.} \\ \midrule[1pt]
        \multirow{7}{*}{\textbf{\makecell{Llama-3.2\\(1B)}}} & 0-Shot & 51.19 & 24.20 & 61.59 & 59.69          & 24.39          & 57.73 & 42.45 & 45.52 & 45.85 \\
                                               & Random    & 62.90 & 28.80 & 80.21 & 90.94          & 42.58          & 52.88 & 43.48 & 42.61 & 55.55 \\
                                               & BM25      & 53.56 & 71.60 & 92.57 & 92.97          & 48.64          & 56.60 & 52.34 & 47.22 & 64.44 \\
                                               & TopK      & 56.84 & 72.80 & 92.78 & 92.53          & 48.82          & 55.67 & 51.22 & 48.38 & 64.88 \\
                                               & MDL       & 59.57 & 82.20 & 92.59 & 93.32          & 48.24          & 56.62 & 52.08 & 49.17 & 66.72 \\
                                               & ConE      & 61.10 & 76.60 & 92.45 & 92.59          & 45.38          & 56.23 & 54.42 & 49.75 & 66.06 \\
         &
          \textbf{D.Va} &
          \textbf{64.46} &
          \textbf{83.00} &
          \textbf{93.30} &
          \textbf{93.52} &
          \textbf{51.63} &
          \textbf{59.95} &
          \textbf{57.61} &
          \textbf{51.10} &
          \textbf{69.32} \\ \midrule[1pt]
        \multirow{7}{*}{\textbf{\makecell{Llama-3.1\\(8B)}}} & 0-Shot & 63.31 & 28.20 & 74.64 & 83.09          & 25.34          & 51.93 & 52.59 & 49.05 & 53.52 \\
                                               & Random    & 73.46 & 38.20 & 84.12 & 96.10          & 45.70          & 56.93 & 67.32 & 57.45 & 64.91 \\
                                               & BM25      & 67.73 & 78.00 & 93.16 & 95.72          & 49.00          & 61.52 & 71.37 & 61.99 & 72.31 \\
                                               & TopK      & 69.21 & 76.40 & 93.28 & 96.05          & 50.09          & 60.92 & 71.08 & 61.50 & 72.31 \\
                                               & MDL       & 71.28 & 86.20 & 93.05 & \textbf{96.96} & \textbf{51.60} & 62.26 & 72.10 & 62.41 & 74.48 \\
                                               & ConE      & 70.60 & 82.80 & 93.33 & 95.55          & 45.20          & 59.91 & 70.83 & 62.72 & 72.62 \\
         &
          \textbf{D.Va} &
          \textbf{73.55} &
          \textbf{86.60} &
          \textbf{93.97} &
          96.43 &
          50.45 &
          \textbf{63.94} &
          \textbf{74.70} &
          \textbf{63.72} &
          \textbf{75.42} \\ \bottomrule
        \end{tabular}%
        }
        \caption{Performance of our method compared to other methods with Llama-3.2-1B and Llama-3.1-8B as the selection and inference model on classification tasks. The best results are highlighted in \textbf{bold}.}
        \label{tab:main-results}
    \end{table*}

    \begin{table*}[h]
        \centering
        \resizebox{0.9\textwidth}{!}{%
        \begin{tabular}{@{}c|c|ccc|ccc|cc|ccc@{}}
        \toprule[1.5pt]
        \multirow{2}{*}{\textbf{Model}} & \multirow{2}{*}{\textbf{Method}} &
          \multicolumn{3}{c|}{\textbf{Flores (de-ru)}} &
          \multicolumn{3}{c|}{\textbf{Flores (en-zh)}} &
          \multicolumn{2}{c|}{\textbf{SQuAD v2}} &
          \multicolumn{3}{c}{\textbf{SamSum}} \\
        & & bleu $\uparrow$ & c20 $\uparrow$ & c22 $\uparrow$ & bleu $\uparrow$ & c20 $\uparrow$ & c22 $\uparrow$ & em $\uparrow$ & f1 $\uparrow$ & r-1 $\uparrow$ & r-2 $\uparrow$ & r-l $\uparrow$ \\ \midrule[1pt]
        \multicolumn{1}{c|}{\multirow{6}{*}{\textbf{\makecell{Llama-3.2\\(1B)}}}} & 0-Shot & 6.69  & -37.49 & 64.77 & 0.49 & -10.01 & 71.43 & 10.65 & 20.34 & 17.60 & 5.86  & 14.87          \\
        \multicolumn{1}{c|}{}                                       & Random    & 9.60  & -2.15  & 72.14 & 1.83 & 20.01  & 77.29 & 19.93 & 27.89 & 36.86 & 14.02 & 28.71          \\
        \multicolumn{1}{c|}{}                                       & BM25      & 9.28  & -4.16  & 71.92 & 5.09 & 19.38  & 77.47 & 19.30 & 27.10 & 38.48 & 15.58 & 30.19          \\
        \multicolumn{1}{c|}{}                                       & TopK      & 9.25  & -0.45  & 72.89 & 3.24 & 20.40  & 77.74 & 20.26 & 28.01 & 39.66 & 16.44 & 31.22          \\
        \multicolumn{1}{c|}{}                                       & ConE      & 9.65  & 2.39   & 73.87 & 2.81 & 22.79  & 77.97 & 17.76 & 26.98 & 40.18 & 16.47 & 31.59          \\
        \multicolumn{1}{c|}{} &
          \textbf{D.Va} &
          \textbf{9.85} &
          \textbf{5.07} &
          \textbf{74.24} &
          \textbf{5.48} &
          \textbf{23.84} &
          \textbf{78.38} &
          \textbf{21.53} &
          \textbf{29.30} &
          \textbf{40.74} &
          \textbf{16.98} &
          \textbf{31.99} \\ \midrule[1pt]
        \multicolumn{1}{c|}{\multirow{6}{*}{\textbf{\makecell{Llama-3.1\\(8B)}}}} & 0-Shot & 17.29 & 44.54  & 82.65 & 1.74 & 41.97  & 82.86 & 20.10 & 30.46 & 4.43  & 1.97  & 3.57           \\
        \multicolumn{1}{c|}{}                                       & Random    & 19.43 & 62.60  & 86.11 & 4.74 & 49.43  & 84.60 & 33.97 & 40.70 & 46.21 & 22.86 & 37.82          \\
        \multicolumn{1}{c|}{}                                       & BM25      & 19.43 & 61.65  & 86.00 & 7.66 & 49.85  & 84.76 & 32.59 & 39.44 & 46.75 & 22.54 & 38.14          \\
        \multicolumn{1}{c|}{}                                       & TopK      & 19.14 & 61.82  & 85.97 & 7.60 & 50.94  & 84.95 & 33.13 & 40.29 & 47.28 & 23.68 & 38.71          \\
        \multicolumn{1}{c|}{}                                       & ConE      & 19.46 & 62.20  & 86.10 & 7.44 & 50.46  & 85.00 & 30.64 & 38.33 & 47.69 & 23.67 & \textbf{39.28} \\
        \multicolumn{1}{c|}{} &
          \textbf{D.Va} &
          \textbf{19.98} &
          \textbf{62.76} &
          \textbf{86.16} &
          \textbf{8.11} &
          \textbf{51.00} &
          \textbf{85.03} &
          \textbf{34.75} &
          \textbf{41.44} &
          \textbf{47.70} &
          \textbf{24.02} &
          38.96 \\ \bottomrule[1.5pt]
        \end{tabular}%
        }
        \caption{Performance of our method compared to other methods with Llama-3.2-1B and Llama-3.1-8B as the selection and inference model on generation tasks. The best results are highlighted in \textbf{bold}. Specifically, c20 and c22 refer to COMET-20 and COMET-22 metrics.}
        \label{tab:nlg-results}
    \end{table*}


    \subsection{Performance with Different Retrieval Models}
    To further demonstrate the effectiveness and robustness of our proposed method, we conduct experiments to evaluate the impact of different retrieval models on the final performance of our method.

    In our experiments, we consider six commonly-used retrieval models, including four models from Sentence-Bert~\citep{reimers-gurevych-2019-sentence}, DPR~\citep{karpukhin-etal-2020-dense} and bge-m3~\citep{bge-m3}. As depicted in Table~\ref{tab:ret-exp} and Table~\ref{tab:full-ret-exp}, our proposed method \textbf{D.Va} consistently outperforms other methods across all retrieval models, demonstrating the effectiveness and robustness of our method under different retrieval models. Moreover, the superior performance of our method is more pronounced when using more powerful retrieval models.

    \begin{table}[t]
        \centering
        \resizebox{0.85\columnwidth}{!}{%
        \begin{tabular}{@{}c|cccc@{}}
            \toprule
            \textbf{Retriever}            & \textbf{TopK}  & \textbf{ConE}  & \textbf{MDL}   & \textbf{D.Va}         \\ \midrule
            \textbf{BM25}     & 64.44 & 65.75 & 66.25 & \textbf{67.83} \\
            \textbf{all-MiniLM-L6-v2}     & 63.55 & 65.50 & 65.64 & \textbf{67.34} \\
            \textbf{all-MiniLM-L12-v2}    & 63.68 & 65.29 & 65.85 & \textbf{67.91} \\
            \textbf{DPR}                  & 63.83 & 65.87 & 65.24 & \textbf{67.77} \\
            \textbf{all-distilroberta-v1} & 63.87 & 65.68 & 65.87 & \textbf{68.57} \\
            \textbf{bge-m3}               & 64.64 & 66.69 & 67.05 & \textbf{69.34} \\
            \textbf{all-mpnet-base-v2}    & 64.88 & 66.06 & 66.72 & \textbf{69.32} \\ \bottomrule
            \end{tabular}%
        }
        \caption{Average performance comparison between \textbf{D.Va} and other methods on different retrieval models. The detailed results are listed in Appendix~\ref{sec:appendix_ret}.
        }
        \label{tab:ret-exp}
    \end{table}

    \begin{table}[h]
    \centering
    \resizebox{0.78\columnwidth}{!}{%
    \begin{tabular}{@{}cc|cc|c@{}}
    \toprule
    \multirow{2}{*}{\textbf{\makecell{Inference\\Model}}} & \multirow{2}{*}{\textbf{Method}} & \multicolumn{2}{c|}{\textbf{Scoring Model}} & \multirow{2}{*}{\textbf{\makecell{Relative\\Change}}} \\
                                           &               & \textit{Itself} & \textit{SLM} &        \\ \midrule
    \multirow{3}{*}{\textbf{GPT2-XL}}      & MDL           & 62.40           & 61.35 & -1.68\%       \\
                                           & ConE          & 61.95           & 61.32                 & -1.02\% \\
                                           & \textbf{D.Va} & \textbf{64.49}           & \textbf{64.73}                 & \textbf{0.37\%}  \\ \midrule
    \multirow{3}{*}{\textbf{\makecell{Llama-3.2\\(3B)}}} & MDL           & 71.27           & 70.91                      & -0.50\%       \\
                                           & ConE          & 70.81           & 70.26                 & -0.78\%  \\
                                           & \textbf{D.Va} & \textbf{72.83}           & \textbf{72.62}                 & \textbf{-0.28\%} \\ \midrule
    \multirow{3}{*}{\textbf{\makecell{Llama-3.1\\(8B)}}} & MDL           & 74.48           & 73.34                 & -1.54\% \\
                                           & ConE          & 72.62           & 72.74                 & \textbf{0.17\%}  \\
                                           & \textbf{D.Va} & \textbf{75.42}           & \textbf{75.20}                 & -0.30\% \\ \bottomrule
    \end{tabular}%
    }
    \caption{Cross-model generalization performance with small language model (\emph{i.e.}, Llama-3.2-1B) as the selection model while other larger language models as the inference model.}
    \label{tab:cross}
    \end{table}

    \subsection{Cross-Model Generalization}
    As our method is model-dependent, we further investigate the generalization ability of our method by selecting and re-ranking with smaller language models while inferring with larger language models. We utilize Llama-3.2-1B as the demonstration selection model, then use larger language models including GPT2-XL, Llama-3.2-3B, and Llama-3.1-8B as the inference model.

    As shown in Table~\ref{tab:cross}, \textbf{D.Va} exhibits strong cross-model generalizability from smaller language models to larger language models, with a relative drop of less than $0.30\%$ on Llama-3.2-3B and Llama-3.1-8B and even performs slightly better than self-selection performance on GPT2-XL. Overall, \textbf{D.Va} demonstrates the most robust cross-model generalization capabilities. Furthermore, even when demonstration selection is performed on the small language model, \textbf{D.Va} significantly outperforms other methods, highlighting its advantage in cost-efficiency.

    \section{Analysis \& Ablation Study}
    \subsection{Impact of In-Context Examples}
    To further substantiate the efficacy of our proposed method, we perform a comparative analysis of \textbf{D.Va} against other methodologies under varying amounts of in-context examples.
    In addition to the 8 demonstrations presented in our primary results, we evaluate the performance of different methods with 1, 2, 4, and 16 demonstrations.


    As depicted in Figure~\ref{fig:in-context-example}, the performance of our proposed method \textbf{D.Va} demonstrates a stable improvement with the increasing number of in-context examples. Notably, \textbf{D.Va} consistently outperforms other methods in all datasets, regardless of the number of demonstrations. In addition, a more detailed comparison of different demonstration selection methods in different numbers of in-context examples is listed in Appendix~\ref{sec:appendix_icl}, results show that our method surpasses all baselines across almost all datasets.

    \subsection{Impact of Coefficient $\lambda$}
\begin{figure}[h]
      \centering
      \begin{subfigure}[b]{0.89\columnwidth}
        \centering
        \includegraphics[width=\columnwidth]{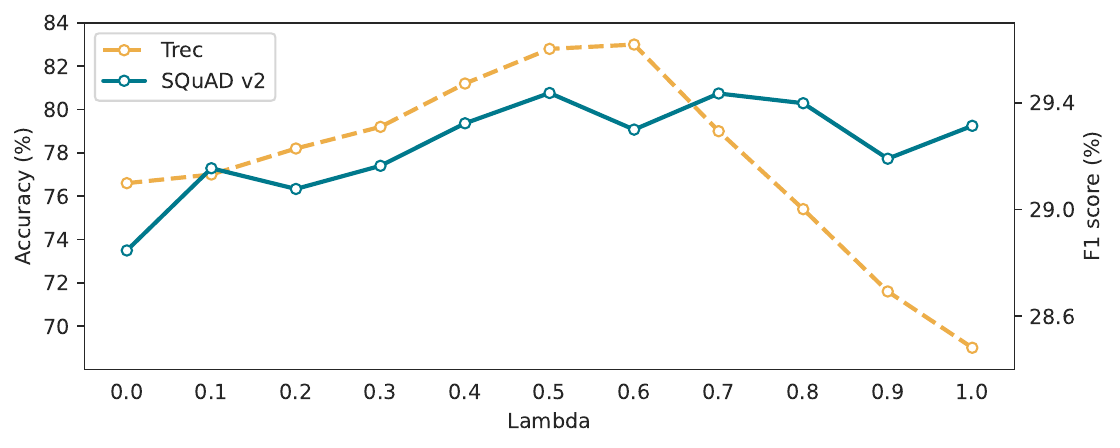}
        \end{subfigure}
    \begin{subfigure}[b]{0.89\columnwidth}
        \centering
        \includegraphics[width=\columnwidth]{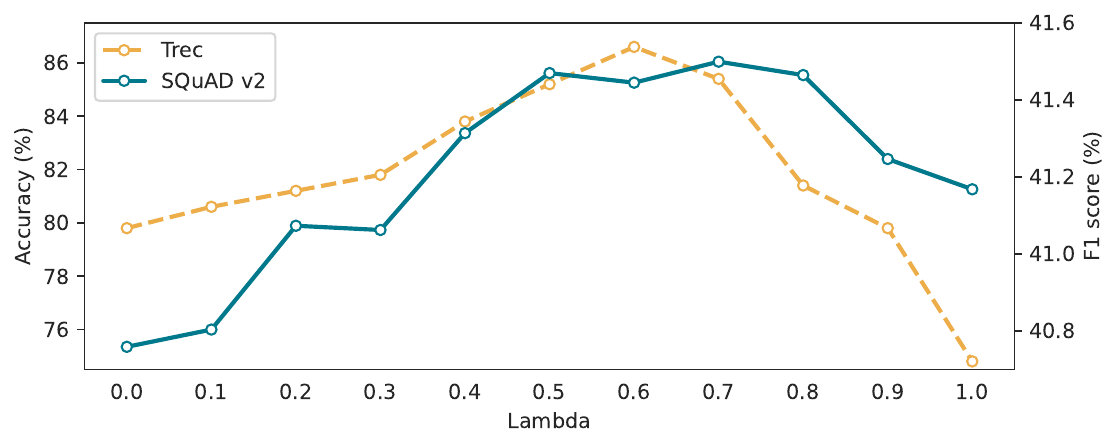}
        \end{subfigure}
      \caption{
      The performance of our method on Trec and SQuAD v2 using different values of $\lambda$ on Llama-3.2-1B (top) and Llama-3.1-8B (bottom). 
      }
      \label{fig:lambda_by_dataset}
    \end{figure}

    For the hyper-parameter $\lambda$, we conduct experiments to investigate the impact of $\lambda$ on the final performance of our method with other hyper-parameters unchanged. We first present the results of one classification dataset Trec and one generation dataset SQuAD v2 in Figure~\ref{fig:lambda_by_dataset}. Experiments are conducted with Llama-3.2-1B and Llama-3.1-8B.

    The results indicate that performance peaks when $\lambda$ is set to around $0.6$ for both Llama-3.2-1B and Llama-3.1-8B. Moreover, the overall performance of our method in eight NLU datasets with varying $\lambda$ in Llama-3.2-1B is shown in Figure~\ref{fig:lambda}.


    The results indicate that the overall performance of our method exhibits a stable trend of initially increasing and then decreasing as the value of $\lambda$ changes, and consistently outperforms all previous methods across a wide range of $\lambda$ values from $0.3$ to $1.0$. Besides, two noteworthy special cases are when $\lambda=0.0$ and $\lambda=1.0$, where the final score is solely determined by the validation loss $\mathcal{L}_{v}$ and the preference-based calibration remainder $\epsilon$, respectively. We further conduct a detailed analysis in Appendix~\ref{sec:lambda_ana}.

    \subsection{Impact of Validation Example Selection}
    \label{sec:valid_example}
    To investigate the influence of validation example selection on the final performance of our method, we conduct experiments utilizing either a randomly chosen example or the furthest example from the retrieved subset as the validation instance. Table~\ref{tab:valid-exp} illustrates the performance of \textbf{D.Va} with three types of validation example selection methods on Llama-3.2-1B. Detailed results are listed in Appendix~\ref{sec:appendix_val}, which demonstrates that the nearest example selection method consistently outperforms the other two methods on most datasets.

    \begin{table}[h]
      \centering
      \resizebox{0.88\columnwidth}{!}{%
      \begin{tabular}{@{}c|ccc@{}}
      \toprule
      \textbf{Validation Example} & \textbf{Random} & \textbf{Furthest} & \textbf{Nearest} \\ \midrule
      \textbf{Avg. Accuracy (\%)} & 67.65           & 67.95             & \textbf{69.32}   \\ \bottomrule
      \end{tabular}%
      }
      \caption{Average performance of our method with different validation example selection methods.}
      \label{tab:valid-exp}
    \end{table}




    

    \subsection{Computational Costs Analysis}
    Computational costs are another critical factor that affects the performance of Ret-ICL methods. Table~\ref{tab:cost-analy} illustrates the overall computational costs of three self-adaptive methods mentioned in this paper.

    \begin{table}[h]
      \centering
      \resizebox{\columnwidth}{!}{%
      \begin{tabular}{@{}c|c|c|c@{}}
        \toprule \textbf{Method} & \textbf{Computational Costs Analysis} & \textbf{Relative Costs} & \textbf{GPU hours} \\ \midrule
        \textbf{MDL} & $\propto$ num. of options \& organizations & $\clubsuit\clubsuit\clubsuit\clubsuit$ & 33 \\
        \textbf{ConE} & $\propto$ num. of candidates $K$ & $\clubsuit$ & 8 \\
        \textbf{D.Va} & $\propto$ num. of candidates $K$ & $\clubsuit\clubsuit$ & 15 \\
        \bottomrule
      \end{tabular}%
      }
      \caption{Computational costs analysis of three self-adaptive methods. \textit{Relative Costs} and \textit{GPU hours} give the relative computational costs and the real-time GPU hours across all datasets under their default settings: the more $\clubsuit$, the higher the real-time costs.}
      \label{tab:cost-analy}
    \end{table}
    
\begin{figure}[h]
      \centering
      \includegraphics[width=\columnwidth]{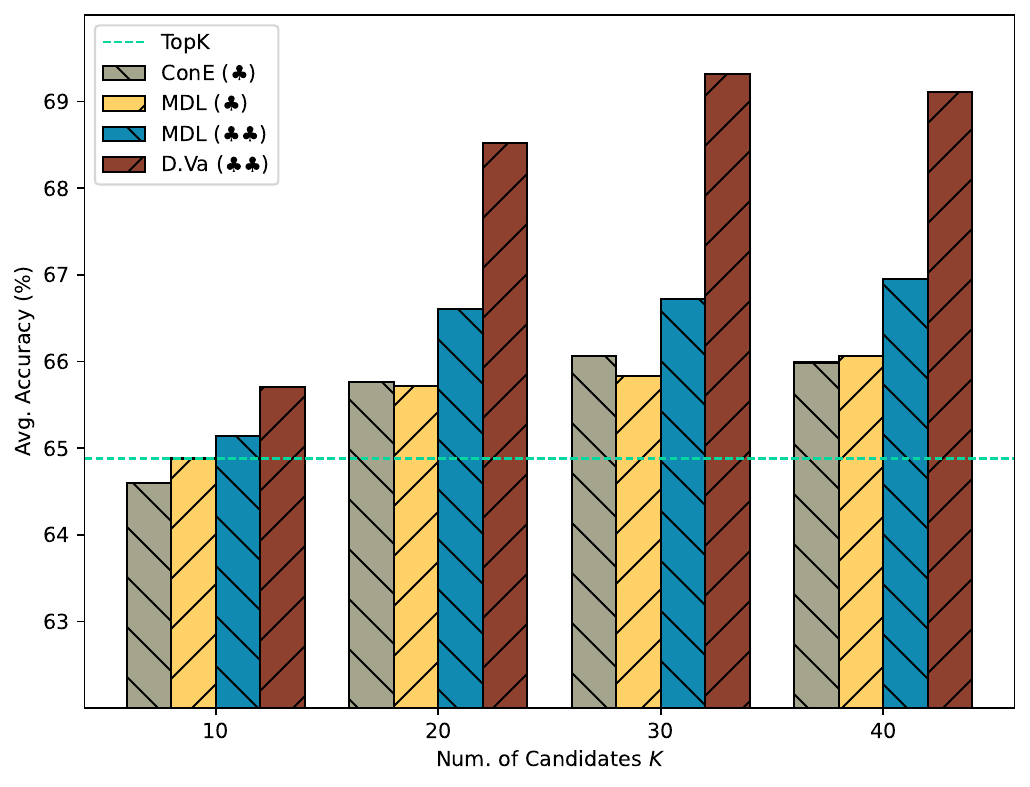}
      \caption{Impact of the number of candidates retrieved by the TopK method. The amount of $\clubsuit$ refers to the real-time costs under the same value of $K$.}
      \label{fig:cand_num}
    \end{figure}

    Considering that the time complexity of our method is positively correlated to the value of $K$, we adjust the overall computational cost by shifting the hyper-parameter $K$. Figure~\ref{fig:cand_num} depicts the performance of our method with different candidate subset sizes compared to MDL and ConE. Under the same $K$, we provide two results of MDL with varying numbers of organizations (discussed in~\citeauthor[Section 6.4]{wu-etal-2023-self}) referring to different overall computational costs.
    
    We can also observe that MDL and ConE perform similarly under the same computational cost, while our method \textbf{D.Va} largely outperforms the MDL method with the same computational cost. By taking MDL as an intermediate reference, we can conclude that our method is also effective when considering the computational costs. Besides, both ConE and \textbf{D.Va} present a performance drop when the value of $K$ continues to grow, which is identical to the conclusion drawn in~\cite{peng-etal-2024-revisiting}.



    \section{Conclusion}
    In this paper, we first introduce a demonstration validation perspective into the self-adaptive demonstration selection process in in-context learning. We propose a novel method \textbf{D.Va} that surpasses all existing retrieval-based in-context learning methods on both natural language understanding (NLU) and natural language generation (NLG) tasks. We further present the generalizability of our methods under different language models, different retrieval models, the number of demonstrations, and numerous datasets.

    \section{Limitations}
    Despite \textbf{D.Va} achieving significant results on the mainstream GPT2-XL and Llama-3 series, due to cost constraints, we have not been able to validate our approach on larger language models.
    Overall, our research empirically showcases the superiority of introducing the demonstration validation mechanism to the demonstration selection field. Although this introduces a minor overhead in the demonstration selection phase, it significantly outperforms previous methods within an acceptable cost margin.


    \bibliography{main}

    \appendix
    \section{Datasets}
    \label{sec:dataset}
    Dataset information is detailed in Table~\ref{tab:dataset}.
    
    \begin{table}[h]
        \centering
        \resizebox{\linewidth}{!}{
        \begin{tabular}{l|ccc}
        \toprule
        \textbf{Dataset} & \textbf{Task} & \textbf{Data Split} \\ \midrule
        \textbf{CMS QA}  &  Commonsense Question Answering & 9471/1221/1140 \\ 
         \textbf{Trec}   &  Topic Classification & 5452/0/500 \\
         \textbf{AgNews} &  Topic Classification & 120000/0/7600 \\
         \textbf{SST-2}  &  Sentiment Classification & 6920/872/1821/ \\
         \textbf{SST-5}  &  Sentiment Classification & 8544/1101/2210 \\
          \textbf{QNLI}  &  Natural Language Inference & 104743/5463/5463 \\
         \textbf{SNLI}   &  Natural Language Inference & 550152/10000/10000 \\
         \textbf{MNLI}   &  Natural Language Inference & 392702/19647/19643 \\
        \bottomrule
        \end{tabular}
        }
        \caption{Details of datasets.}
        \label{tab:dataset}
    \end{table}

    \section{Evaluation Strategy}
    \label{sec:eval}

    For the classification tasks, we evaluate the performance of our method on the test set using the accuracy metric.
    For the generation tasks, we evaluate the performance of our method on the test set using the exact match score and f1 score for the SQuAD v2 dataset and the ROUGE score for the Samsum dataset.
    Specifically, we follow the model-based evaluation settings of \citet{peng-etal-2024-revisiting}, using COMET metrics\footnote{The score of COMET-20 metric is unbounded but typically falls between -1 and 1 where 1 reflects a perfect translation, while the score of COMET-22 metric is between 0 and 1 where 1 represents a perfect translation. We have expanded these two indicators by a factor of 100 to more clearly distinguish between the superior and inferior.}~\citep{rei-etal-2020-comet,rei-etal-2022-comet} for the translation tasks and the corresponding BLEU scores are also considered. Besides, we use a 1-shot setting for SQuAD V2 and a 3-shot setting for the translation tasks.
    
    \section{Performance with Different Retrieval Models}
    \label{sec:appendix_ret}
    The detailed performance of our proposed \textbf{D.Va} compared to previous methods with different retrieval models are shown in Table~\ref{tab:full-ret-exp}. Our \textbf{D.Va} surpasses all the methods in almost all datasets. Furthermore, as the capability of the retrieval model increases, the performance advantage of \textbf{D.Va} becomes more pronounced. We hypothesize that this may be related to \textbf{D.Va}'s relative reliance on the optimal validation example.

    \section{Analysis \& Ablation Study}
    \subsection{Impact of Validation Example Selection}
    \label{sec:appendix_val}
    The detailed performance of \textbf{D.Va} with three different validation example selection methods are listed below. As depicted in~\ref{tab:full-valid-exp}, selecting the semantically nearest demonstration as the validation example consistently outperforms the other two selecting methods, especially for Trec dataset and Commonsense QA dataset.
    \begin{table}[h]
    \centering
    \resizebox{0.9\columnwidth}{!}{%
    \begin{tabular}{@{}c|ccc@{}}
    \toprule
    \textbf{Valid. Example} & \textbf{Random} & \textbf{Furthest} & \textbf{Nearest} \\ \midrule
    \textbf{Trec}           & 75.00           & 76.80             & \textbf{83.00}   \\
    \textbf{CMS QA}         & 61.51           & 59.79             & \textbf{64.46}   \\
    \textbf{SST-2}          & 92.86           & 93.30             & \textbf{93.52}   \\
    \textbf{SST-5}          & 50.36           & 51.54             & \textbf{51.63}   \\
    \textbf{QNLI}           & 60.48           & \textbf{60.94}    & 59.95            \\
    \textbf{AgNews}         & \textbf{93.58}  & 93.42             & 93.30            \\
    \textbf{SNLI}           & 57.17           & \textbf{57.83}    & 57.61            \\
    \textbf{MNLI}           & 50.25           & 49.98             & \textbf{51.10}   \\ \midrule
    \textbf{Avg.}           & 67.65           & 67.95             & \textbf{69.32}   \\ \bottomrule
    \end{tabular}%
    }
    \caption{Results of different validation example selection methods.}
    \label{tab:full-valid-exp}
    \end{table}

    \subsection{Impact of In-Context Examples}
    \label{sec:appendix_icl}
\begin{figure*}[h]
      \centering
      \includegraphics[width=\textwidth]{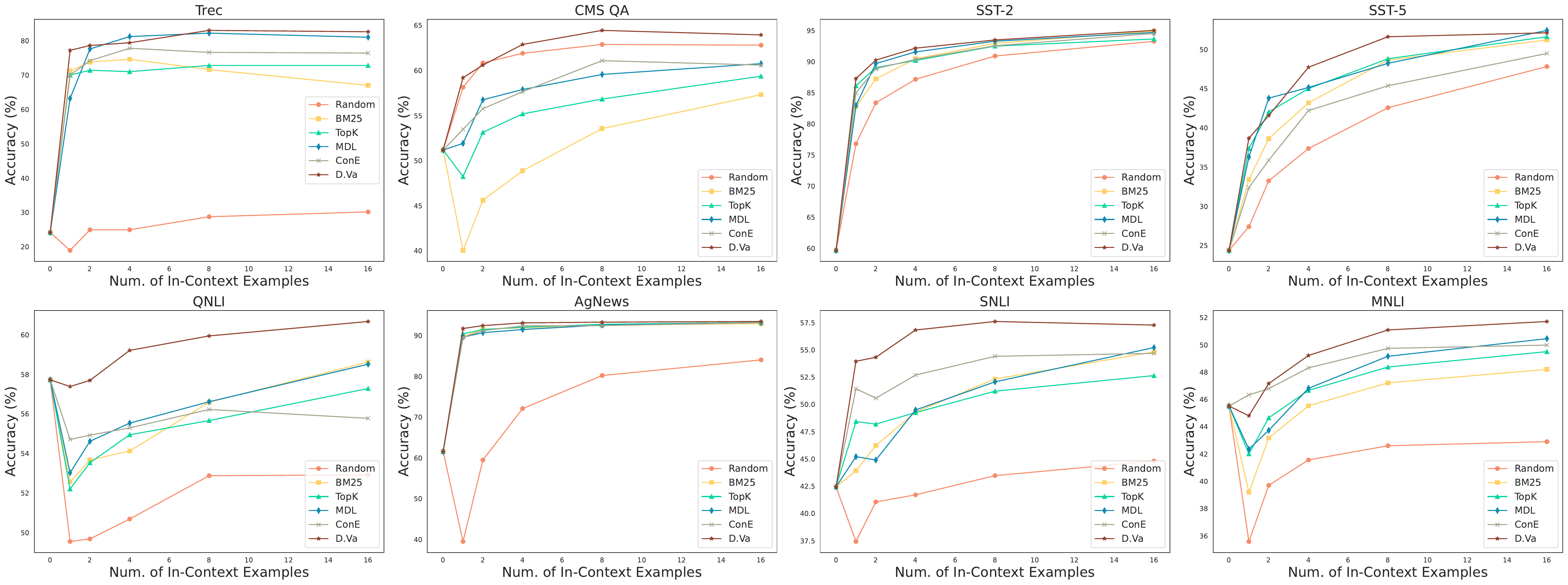}
      \caption{Impact of the number of in-context examples on different datasets on Llama-3.2-1B.}
      \label{fig:all-icl-num}
    \end{figure*}

    In this section, we provide a detailed comparison of the performance of \textbf{D.Va} and previous methods across different amount of in-context examples in Figure~\ref{fig:all-icl-num}. As the number of in-context examples increases, the performance of all methods improves. Notably, \textbf{D.Va} maintains a consistent lead.

    \subsection{Demonstration Ordering}
    \label{sec:demons_order}
    In this section, we investigate the impact of demonstration ordering on the final performance of our method. We conduct experiments to evaluate the performance of our method with three intuitive ordering methods, including the descending order (ours), the ascending order, and the randomly shuffled order. In particular, we conducted experiments with three random seeds for the third ordering method and reported the average performance as well as the standard deviation.

    \begin{table}[h]
      \centering
      \resizebox{0.9\columnwidth}{!}{%
      \begin{tabular}{@{}c|ccc@{}}
      \toprule
      \textbf{Ordering}       & \textbf{Ascending} & \textbf{Random} & \textbf{Descending} \\ \midrule
      \textbf{Llama-3.2-1B} & 68.68 & 68.89$\pm$ 0.19 & \textbf{69.32} \\
      \textbf{Llama-3.1-8B} & 75.19 & 75.27$\pm$ 0.08 & \textbf{75.42} \\ \bottomrule
      \end{tabular}%
      }
      \caption{The impact of ordering methods on the final performance of our method.}
      \label{tab:order-exp}
    \end{table}

    Table~\ref{tab:order-exp} depicted that the performance of our method with the descending order of demonstrations outperforms the other two ordering methods, indicating that learning from the less informative demonstrations first can help the language model better understand the test input. Besides, despite the poor performance compared to the other two ordering methods, the ascending ordering under \textbf{D.Va} selection still outperforms other methods, demonstrating the effectiveness of our method.
    
    \begin{table*}[h]
    \centering
    \resizebox{\textwidth}{!}{%
    \begin{tabular}{@{}cc|cccccccc|c@{}}
    \toprule[1.5pt]
    \multicolumn{1}{c|}{\textbf{Retriever}} &
      \textbf{Methods} &
      \textbf{CMS QA} &
      \textbf{Trec} &
      \textbf{AgNews} &
      \textbf{SST-2} &
      \textbf{SST-5} &
      \textbf{QNLI} &
      \textbf{SNLI} &
      \textbf{MNLI} &
      \textbf{Avg.} \\ \midrule[1pt]
    \multicolumn{1}{c|}{/} &
      0-Shot &
      51.19 &
      24.20 &
      61.59 &
      59.69 &
      24.39 &
      57.73 &
      42.45 &
      45.52 &
      45.85 \\ \midrule[1pt]
    \multicolumn{1}{c|}{/} &
      Random &
      62.90 &
      28.80 &
      80.21 &
      90.94 &
      42.58 &
      52.88 &
      43.48 &
      42.61 &
      55.55 \\ \midrule[1pt]
    \multicolumn{1}{c|}{\multirow{4}{*}{\textbf{BM25}}} &
      BM25 &
      53.56 &
      71.60 &
      92.57 &
      92.97 &
      \textbf{48.64} &
      56.60 &
      52.34 &
      47.22 &
      64.44 \\
    \multicolumn{1}{c|}{} &
      MDL &
      56.51 &
      \textbf{81.00} &
      92.67 &
      \textbf{94.07} &
      48.10 &
      57.06 &
      53.04 &
      47.56 &
      66.25 \\
    \multicolumn{1}{c|}{} &
      ConE &
      57.66 &
      78.20 &
      92.75 &
      92.75 &
      43.53 &
      56.32 &
      56.15 &
      48.63 &
      65.75 \\
    \multicolumn{1}{c|}{} &
      \textbf{D.Va} &
      \textbf{62.16} &
      80.40 &
      \textbf{92.93} &
      \textbf{94.07} &
      47.83 &
      \textbf{58.74} &
      \textbf{57.61} &
      \textbf{48.87} &
      \textbf{67.83} \\ \midrule[1pt]
    \multicolumn{1}{c|}{\multirow{4}{*}{\textbf{all-MiniLM-L6-v2}}} &
      TopK &
      56.27 &
      68.20 &
      92.67 &
      91.10 &
      47.42 &
      54.99 &
      50.41 &
      47.31 &
      63.55 \\
    \multicolumn{1}{c|}{} &
      MDL &
      58.07 &
      \textbf{79.60} &
      92.80 &
      \textbf{92.75} &
      46.47 &
      56.16 &
      50.99 &
      48.28 &
      65.64 \\
    \multicolumn{1}{c|}{} &
      ConE &
      58.31 &
      78.00 &
      92.99 &
      92.26 &
      44.98 &
      55.46 &
      53.08 &
      \textbf{48.91} &
      65.50 \\
    \multicolumn{1}{c|}{} &
      \textbf{D.Va} &
      \textbf{62.00} &
      79.00 &
      \textbf{93.08} &
      92.70 &
      \textbf{48.01} &
      \textbf{59.67} &
      \textbf{55.47} &
      48.77 &
      \textbf{67.34} \\ \midrule[1pt]
    \multicolumn{1}{c|}{\multirow{4}{*}{\textbf{all-MiniLM-L12-v2}}} &
      TopK &
      55.61 &
      67.60 &
      92.64 &
      91.93 &
      48.14 &
      54.68 &
      50.74 &
      48.12 &
      63.68 \\
    \multicolumn{1}{c|}{} &
      MDL &
      57.66 &
      \textbf{79.60} &
      92.22 &
      92.15 &
      48.14 &
      56.67 &
      51.48 &
      48.89 &
      65.85 \\
    \multicolumn{1}{c|}{} &
      ConE &
      57.66 &
      76.60 &
      92.63 &
      92.70 &
      44.48 &
      55.13 &
      53.49 &
      49.64 &
      65.29 \\
    \multicolumn{1}{c|}{} &
      \textbf{D.Va} &
      \textbf{63.06} &
      78.40 &
      \textbf{92.97} &
      \textbf{93.63} &
      \textbf{48.64} &
      \textbf{59.60} &
      \textbf{56.74} &
      \textbf{50.27} &
      \textbf{67.91} \\ \midrule[1pt]
    \multicolumn{1}{c|}{\multirow{4}{*}{\textbf{DPR}}} &
      TopK &
      56.02 &
      66.80 &
      91.89 &
      92.64 &
      47.96 &
      54.73 &
      53.31 &
      47.28 &
      63.83 \\
    \multicolumn{1}{c|}{} &
      MDL &
      56.76 &
      74.20 &
      91.66 &
      \textbf{93.08} &
      46.97 &
      56.10 &
      55.42 &
      47.75 &
      65.24 \\
    \multicolumn{1}{c|}{} &
      ConE &
      58.72 &
      77.00 &
      92.53 &
      92.42 &
      44.39 &
      56.34 &
      56.92 &
      48.67 &
      65.87 \\
    \multicolumn{1}{c|}{} &
      \textbf{D.Va} &
      \textbf{62.08} &
      \textbf{78.60} &
      \textbf{92.83} &
      92.70 &
      \textbf{48.28} &
      \textbf{59.95} &
      \textbf{58.56} &
      \textbf{49.20} &
      \textbf{67.77} \\ \midrule[1pt]
    \multicolumn{1}{c|}{\multirow{4}{*}{\textbf{all-distilroberta-v1}}} &
      TopK &
      58.48 &
      64.40 &
      92.54 &
      92.75 &
      48.28 &
      54.84 &
      51.43 &
      48.26 &
      63.87 \\
    \multicolumn{1}{c|}{} &
      MDL &
      58.80 &
      78.20 &
      92.29 &
      93.14 &
      47.19 &
      56.20 &
      52.30 &
      48.87 &
      65.87 \\
    \multicolumn{1}{c|}{} &
      ConE &
      58.15 &
      76.40 &
      92.55 &
      92.53 &
      46.65 &
      55.04 &
      54.63 &
      49.45 &
      65.68 \\
    \multicolumn{1}{c|}{} &
      \textbf{D.Va} &
      \textbf{63.55} &
      \textbf{80.20} &
      \textbf{93.20} &
      \textbf{93.79} &
      \textbf{50.00} &
      \textbf{60.88} &
      \textbf{56.59} &
      \textbf{50.32} &
      \textbf{68.57} \\ \midrule[1pt]
    \multicolumn{1}{c|}{\multirow{4}{*}{\textbf{bge-m3}}} &
      TopK &
      55.77 &
      69.80 &
      92.82 &
      91.65 &
      49.59 &
      56.69 &
      50.15 &
      50.64 &
      64.64 \\
    \multicolumn{1}{c|}{} &
      MDL &
      59.38 &
      80.60 &
      92.59 &
      92.70 &
      48.78 &
      59.00 &
      51.85 &
      51.48 &
      67.05 \\
    \multicolumn{1}{c|}{} &
      ConE &
      58.31 &
      80.00 &
      93.07 &
      92.48 &
      47.15 &
      57.57 &
      54.38 &
      50.59 &
      66.69 \\
    \multicolumn{1}{c|}{} &
      \textbf{D.Va} &
      \textbf{64.62} &
      \textbf{82.40} &
      \textbf{93.58} &
      \textbf{94.01} &
      \textbf{49.64} &
      \textbf{61.52} &
      \textbf{56.02} &
      \textbf{52.90} &
      \textbf{69.34} \\ \midrule[1pt]
    \multicolumn{1}{c|}{\multirow{4}{*}{\textbf{all-mpnet-base-v2}}} &
      TopK &
      56.84 &
      72.80 &
      92.78 &
      92.53 &
      48.82 &
      55.67 &
      51.22 &
      48.38 &
      64.88 \\
    \multicolumn{1}{c|}{} &
      MDL &
      59.57 &
      82.20 &
      92.59 &
      93.32 &
      48.24 &
      56.62 &
      52.08 &
      49.17 &
      66.72 \\
    \multicolumn{1}{c|}{} &
      ConE &
      61.10 &
      76.60 &
      92.45 &
      92.59 &
      45.38 &
      56.23 &
      54.42 &
      49.75 &
      66.06 \\
    \multicolumn{1}{c|}{} &
      \textbf{D.Va} &
      \textbf{64.46} &
      \textbf{83.00} &
      \textbf{93.30} &
      \textbf{93.52} &
      \textbf{51.63} &
      \textbf{59.95} &
      \textbf{57.61} &
      \textbf{51.10} &
      \textbf{69.32} \\ \bottomrule[1.5pt]
    \end{tabular}%
    }
    \caption{Full results of our method compared to other methods on different retrieval models with Llama-3.2-1B as the selection and inference model on classification tasks. The best results under the same retriever are highlighted in \textbf{bold}.}
    \label{tab:full-ret-exp}
    \end{table*}

    \subsection{Impact of Coefficient $\lambda$}\
    \label{sec:lambda_ana}
    In this section, we further analyze the two special cases of the value of $\lambda$.
    
    \paragraph{When $\lambda$ is set to $0.0$.} For the former case, the performance of which is significantly inferior to other methods due to the distribution shift between the validation example and the test example. This phenomenon is soon alleviated as $\lambda$ increases to $0.3$ and beyond, where the preference-based calibration remainder plays a more significant role in the final score.

    \paragraph{When $\lambda$ is set to $1.0$.} For the latter, the average performance across eight NLU datasets is slightly higher than other methods but still inferior to the optimal performance achieved with $\lambda=0.6$. In this case, demonstrations with a lower preference-based calibration remainder are more likely to be selected, which indicates that the language model prefers the test example over the validation example as the next input after in-context training on the current under-evaluated training example. On the other hand, despite the semantical similarity between the test example and the validation demonstration, the language model still exhibits a preference for the test sample input. Consequently, under this condition, even when only considering the calibration remainder, the performance of our method is preserved to some extent.

    \section{Templates}
    \label{sec:appendix_templates}
    \subsection{Templates for Classification Tasks}
    \begin{table*}[t]
    \centering
    \resizebox{0.9\linewidth}{!}{
    \begin{tabular}{lll}
    \toprule
    \textbf{Dataset} & \textbf{Prompt} & \textbf{Class} \\
    \hline
    \multirow{2}{*}{SST-2} & Review: <X> Sentiment: positive  & Positive \\ 
    & Review: <X> Sentiment: negative  & Negative \\
    \midrule
    \multirow{5}{*}{SST-5}
    & Review: <X> Sentiment: terrible  & Very Negative \\ 
    & Review: <X> Sentiment: bad  & Negative \\
    & Review: <X> Sentiment: okay  & Neutral \\ 
    & Review: <X> Sentiment: good  & Positive \\
    & Review: <X> Sentiment: great  & Very Positive \\ 
    \midrule
    \multirow{3}{*}{SNLI\, \& \, MNLI}
    & <C> Can we know <X>? Yes.  & Entailment \\ 
    & <C> Can we know <X>? Maybe.  & Neutral \\
    & <C> Can we know <X>? No.  & Contradiction \\ 
    \midrule
    
    \multirow{2}{*}{QNLI}
    & <C> Can we know <X>? Yes.  & Entailment \\ 
    & <C> Can we know <X>? No.  & Contradiction \\ 
    \midrule
    \multirow{6}{*}{TREC}
    & "<X>" It is about abbreviation.  & ABBR \\ 
    & "<X>" It is about entity.  & ENTY \\ 
    & "<X>" It is about description and abstract concept.  & DESC \\ 
    & "<X>" It is about human being.  & HUM \\ 
    & "<X>" It is about location.  & LOC \\ 
    & "<X>" It is about numeric value.  & NUM \\ 
    
    \midrule
    
    \multirow{4}{*}{AgNews}
    & Input: <X> Type: world & World \\ 
    & Input: <X> Type: sports  & Sports \\ 
    & Input: <X> Type: business  & Business \\ 
    & Input: <X> Type: technology  & Sci/Tech \\ 
    \midrule
    \multirow{5}{*}{Commonsense QA}
    & Answer the following question:\,<X>\,  Answer: <A>.  & A \\ 
    & Answer the following question:\,<X>\,  Answer: <B>.  & B \\ 
    & Answer the following question:\,<X>\,  Answer: <C>.  & C \\ 
    & Answer the following question:\,<X>\,  Answer: <D>.  & D \\ 
    & Answer the following question:\,<X>\,  Answer: <E>.  & E \\ 
    \bottomrule
    
    \end{tabular}
    }
    \caption{Templates of classification tasks. Placeholders (e.g., <X> and <A>) will be replaced by real inputs or answers (in Commonsense QA).}
    \label{tab:cls_temp}
    \end{table*}
    
    The templates for classification tasks used in this paper are detailed in Table~\ref{tab:cls_temp}.
    \subsection{Templates for Generation Tasks}
    \begin{table*}[]
    \centering
    \resizebox{\textwidth}{!}{%
    \begin{tabular}{@{}ll@{}}
    \toprule
    \textbf{Dataset} & \textbf{Prompt}                                                                                                                 \\ \midrule
    \textbf{Flores}  & \begin{tabular}[c]{@{}l@{}}Translate from <src> to <tgt>:\\ <src>: <source> <tgt>: <target>\end{tabular}                        \\ \midrule
    \textbf{SQuAD v2} &
      \begin{tabular}[c]{@{}l@{}}Answer each question using information in the preceding background paragraph. If there is not enough information provided, answer with "Not in background".\\ \\ Title: <title>\\ \\ \\ Background: <context>\\ \\ \\ Q: <question>\\ \\ \\ A: <answer>\end{tabular} \\ \midrule
    \textbf{SamSum}  & \begin{tabular}[c]{@{}l@{}}What is a summary of this dialogue?\\ Dialogue:\\ <dialogue>\\ Summary: <summary>\end{tabular} \\ \bottomrule
    \end{tabular}%
    }
    \caption{Templates of generation tasks. For Flores, <src> and <tgt> refer to the source and target language. For SQuAD v2, we use a similar format as \href {https://huggingface.co/datasets/meta-llama/Llama-3.2-1B-evals}{Llama-3's evaluation}.}
    \label{tab:gen_temp}
    \end{table*}

    The templates for generation tasks used in this paper are detailed in Table~\ref{tab:gen_temp}.
\end{document}